\documentclass[runningheads]{llncs}
\usepackage[utf8]{inputenc}
\usepackage{color}
\usepackage{graphicx,caption,subcaption,amsmath,amssymb,tabularx,array}
\usepackage{setspace}
\usepackage{booktabs,wrapfig}
\usepackage[T1]{fontenc}
\usepackage{xcolor} 
\definecolor{diffblue}{RGB}{25, 25, 112}
\definecolor{diffgreen}{RGB}{0, 100, 0}

\usepackage{caption}
\PassOptionsToPackage{hyphens}{url}\usepackage[colorlinks=false, pdfborder={0 0 0}]{hyperref}
\title{\textit{SimplifyMyText}: An LLM-Based System for Inclusive Plain Language Text Simplification}

\titlerunning{An LLM-Based System for Inclusive Plain Language Text Simplification}
\author{Michael Färber\textsuperscript{\orcid{0000-0001-5458-8645}} \and 
Parisa Aghdam\textsuperscript{\orcid{0009-0004-3393-1298}} \and
Kyuri Im\textsuperscript{\orcid{0009-0002-1039-4405}} \and
Mario Tawfelis\textsuperscript{\orcid{0009-0006-8745-1737}}\and
Hardik Ghoshal\textsuperscript{\orcid{0009-0005-6747-9559}}}

\authorrunning{M. Färber et al.}

\newcommand{\orcid}[1]{\href{https://orcid.org/#1}{\includegraphics[width=10pt]{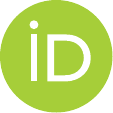}}}

\institute{Technical University (TU) Dresden \& ScaDS.AI, Dresden, Germany 
\email{michael.faerber@tu-dresden.de} 
\email{\{parisa.aghdam,kyuri.im,mario.tawfelis,hardik.ghoshal\}@mailbox.tu-dresden.de} 
}

\begin{document}
\maketitle              %
\begin{abstract}
Text simplification is essential for making complex content accessible to diverse audiences who face comprehension challenges. Yet, the limited availability of simplified materials creates significant barriers to personal and professional growth and hinders social inclusion. Although researchers have explored various methods for automatic text simplification, none fully leverage large language models (LLMs) to offer tailored customization for different target groups and varying levels of simplicity. Moreover, despite its proven benefits for both consumers and organizations, the well-established practice of \textit{plain language} remains underutilized.
In this paper, we introduce \url{https://simplifymytext.org}, the first system designed to produce \textit{plain language} content from multiple input formats, including typed text and file uploads, with flexible customization options for diverse audiences. We employ GPT-4 and Llama-3 and evaluate outputs across multiple metrics. 
Overall, our work contributes to research on automatic text simplification and highlights the importance of tailored communication in promoting inclusivity. 

\keywords{Text Simplification  \and Plain Language \and Large Language Model \and Demo System}
\end{abstract}

\section{Introduction}

Effective communication is crucial for knowledge sharing and meaningful interaction, yet many people struggle to understand written information. In Germany alone, 10 to 17 million individuals face reading challenges \cite{zeit}, including functional illiterates, people with reading and writing disorders, mental disabilities, and refugees. Across Europe, 20-25\% of the population is functionally illiterate \cite{/content/paper/5jm0v44bnmnx-en}, and globally, 16\% of adults—about 759 million—lack basic literacy skills \cite{unEducationAll}. Accessible language is essential for empowering these individuals to understand crucial information.

Text simplification (see Table 1) tackles this issue by producing clearer versions of texts while preserving meaning \cite{siddharthan2014}. Plain language \cite{ISO24495}, a widely recognized approach, aims to make information accessible to broader audiences, benefiting not only those with literacy challenges. Originally developed to help the public comprehend expert content, plain language has been recommended since the 1960s for individuals with communication difficulties \cite{9783732906918}. Its application spans sectors such as government, healthcare, law, and business.
In the U.S., the Plain Writing Act mandates that federal agencies ensure essential documents are accessible, though thorough simplification remains inconsistent. Similarly, the European Union promotes plain language in public documents, but implementation varies across member states \cite{plain-lang-global}.

\begin{table}[tb]
\centering
\caption{Example of Text Simplification}
\label{tab:examples}
\setlength{\tabcolsep}{10pt} %
\resizebox{0.98\textwidth}{!}{%
\begin{tabular}{p{4cm} p{10.5cm}}
\toprule
\textbf{Original Text (paper abstract of \cite{DBLP:conf/nips/KrizhevskySH12}):} & We trained a \textcolor{diffblue}{deep convolutional neural network} to classify \textcolor{diffblue}{1.2 million high-resolution images} from the \textcolor{diffblue}{ImageNet LSVRC-2010 contest} into \textcolor{diffblue}{1000 classes}. We achieved \textcolor{diffblue}{top-1 and top-5 error rates of 37.5\% and 17.0\%}, improving on the \textcolor{diffblue}{previous state-of-the-art}. The network has \textcolor{diffblue}{60 million parameters, 650,000 neurons}, \textcolor{diffblue}{five convolutional layers (some followed by max-pooling)}, and \textcolor{diffblue}{three fully-connected layers} with a final \textcolor{diffblue}{1000-way softmax}. To \textcolor{diffblue}{accelerate training}, we used \textcolor{diffblue}{non-saturating neurons and GPU-optimized convolution}. Overfitting in \textcolor{diffblue}{fully-connected layers} was reduced with \textcolor{diffblue}{‘dropout’ regularization}. \\
\midrule
\textbf{Simplified Text (chosen target audience: \textit{General Public}):} & We used a \textcolor{diffgreen}{complex computer program} to \textcolor{diffgreen}{sort 1.2 million high-quality pictures from a competition} into \textcolor{diffgreen}{1000 categories}. We \textcolor{diffgreen}{improved on previous best results}, with our program \textcolor{diffgreen}{making a mistake only 37.5\% of the top time and 17.0\% of the top five times}. The program has a \textcolor{diffgreen}{complicated structure including 60 million parameters and 650,000 processing units}. To \textcolor{diffgreen}{speed up the process}, we used a \textcolor{diffgreen}{method that avoids slowing down the system} and \textcolor{diffgreen}{another that is optimized for a specific type of computer hardware}. To \textcolor{diffgreen}{avoid the system from being too specific and not generalizing well}, we used a technique called \textcolor{diffgreen}{‘dropout’ to keep things balanced}. \\
\bottomrule
\end{tabular}
}
\end{table}

Manual simplification of content is labor-intensive. Early work by Maass et al. \cite{maass2014} highlighted the potential of computer-aided tools for translating content into \textit{easy language},\footnote{\textit{Plain language} aims to make content understandable to a general audience, while \textit{easy language} is a more simplified form tailored specifically for individuals with cognitive or reading disabilities \cite{9783732906918}.} and subsequent studies \cite{sheang-saggion-2021-controllable,maddela-etal-2021-controllable,martin-etal-2020-controllable} advanced automatic text simplification using deep learning. More recent approaches leverage pre-trained models \cite{omelianchuk-etal-2021-text,devaraj-etal-2022-evaluating}, with large language models (LLMs) emerging as efficient and adaptable alternatives \cite{anschutz-etal-2023-language}.
LLMs bring significant flexibility to natural language processing tasks through zero-shot and few-shot learning \cite{brown2020language,thoppilan2022lamda}. Studies by Swanson et al. \cite{Swanson2024} demonstrate that fine-tuned LLMs excel in simplifying complex biomedical information, and Araújo et al. \cite{araujo2023simplifying} developed a ChatGPT-based method to enhance interpretability for non-experts. However, applying LLMs specifically for generating plain language remains underexplored, and no existing system fully meets the need for an interactive platform that both generates plain language and customizes output for specific audiences.

In this paper, we introduce \url{https://simplifymytext.org}, an online LLM-based system that translates text into plain language tailored for various audiences. As the first system of its kind, it enables professionals to refine outputs for specific audiences. With a user-friendly design, the platform lets users generate customized plain language texts by pasting content or uploading documents, producing versions aligned with their intended audience.

\begin{figure}[tb]
    \centering
    \includegraphics[width=0.99\textwidth]{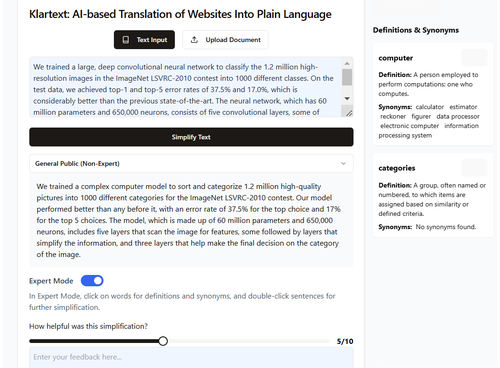} %
    \caption{Screenshot of our demo system at \url{https://simplifymytext.org}}
    \label{fig:my_label}
\end{figure}

\section{Demonstration System}

Our code and demonstration video are available on GitHub,\footnote{\url{https://github.com/grussdorian/klartext}} and the service can be accessed online.\footnote{\url{https://simplifymytext.org}} A screenshot of the platform is shown in Fig.~\ref{fig:my_label}. Built with React \cite{de2020humanportal} for the front end and Node.js for the back end, the interface streamlines text submission in various formats, including typed input, PDFs, and Word documents. By clicking “Simplify Text,” users activate an LLM-based approach (defaulting to OpenAI’s \texttt{gpt-4o} model). This approach leverages zero-shot learning, enabling text simplification without requiring task-specific training. 
It is designed for flexibility, supporting the integration of any large language model to promote open-source adaptability and institutional accessibility.

The service offers five audience-specific simplification options:
(1)~\textit{Scientists and Researchers},
(2)~\textit{Students and Academics},
(3)~\textit{Industry Professionals},
(4)~\textit{Journalists and Media Professionals}, and
(5)~\textit{General Public/Non-Experts}.
If no audience is specified, simplification defaults to the \textit{General Public}. 

An “Expert Mode” provides advanced editing tools for further refinement. Users can click on words to access synonyms or definitions and select sentences to view rephrased versions at varying complexity levels, enabling precise adjustments for comprehension. A rating feature creates a feedback loop to continually enhance the tool’s effectiveness and accuracy. This combination of audience-specific customization, adjustable simplification, and expert editing options establishes the platform as a purpose-built alternative to generic AI text tools.

\section{Evaluation} 

We evaluate our system using the PKWP dataset~\cite{10.5555/1873781.1873933}, which contains 108,016 paired sentences from 65,133 Wikipedia articles. To assess performance and optimize prompts, we use three metrics: \textit{BLEU}, \textit{Flesch-Kincaid}, and \textit{SARI}. \textit{Bilingual Evaluation Understudy (BLEU)} \cite{Papineni2002} measures similarity between the simplified text and human references, with scores ranging from 0 to 1, where 1 indicates perfect alignment. The \textit{Flesch Reading Ease (FRE)}~\cite{10.5555/1873781.1873933} score gauges readability based on sentence and word length, while \textit{System Output Against References and Against the Input Sentence (SARI)} \cite{xu-etal-2016-optimizing} evaluates simplification quality by examining how effectively the simplified output adds, deletes, and retains words compared to the original text and human references. Higher scores on these metrics indicate improved readability and closer alignment with accessibility standards.

\begin{table}[tb]
\centering
\caption{Evaluation across different user groups (highest values in bold, second-highest underlined).}
\label{tab:user-group-metrics}
\begin{small}
\begin{tabular}{l@{\hspace{0.3cm}}l@{\hspace{0.2cm}}r@{\hspace{0.2cm}}r@{\hspace{0.2cm}}r@{\hspace{0.2cm}}r}
\toprule
\textbf{User Group}             & \textbf{Model} & \textbf{BLEU} & \textbf{SARI}  & \textbf{FK Ease} & \textbf{FK Grade} \\
\midrule
Scientists and         & GPT-4o   & 0.434 & 39.78 & 74.49   & 6.30     \\ 
Researchers            & Llama 3.1 & 0.485 & \underline{40.68} & 69.11   & 8.30     \\ \midrule
Students and           & GPT-4o   & 0.443 & 40.37 & \textbf{75.20}   & \textbf{6.00}     \\ 
Academics              & Llama 3.1 & \textbf{0.509} & \textbf{41.27} & 68.81   & 8.50     \\ \midrule
Industry Professionals & GPT-4o   & 0.423 & 38.97 & 66.03   & 7.50     \\ 
                       & Llama 3.1 & \underline{0.486} & 40.67 & 69.31   & 8.30     \\ \midrule
Journalists and        & GPT-4o   & 0.414 & 39.13 & \underline{74.79}   & \underline{6.20}     \\ 
Media                  & Llama 3.1 & 0.471 & 40.53 & 69.62   & 8.10     \\ \midrule
General Public         & GPT-4o   & 0.439 & 40.15 & \underline{74.79}   & \underline{6.20}     \\ 
                       & Llama 3.1 & 0.446 & 39.93 & 69.62   & 8.10     \\
\bottomrule
\end{tabular}%
\end{small}
\end{table}

Table \ref{tab:user-group-metrics} presents the evaluation results generated by GPT-4o, a model with approximately 1.8 trillion parameters, and LLaMA 3.1, a model with 70 billion parameters, across user groups. \textit{BLEU} scores are consistent in both models, indicating a good alignment with reference texts in all groups.
The \textit{SARI} scores are highest for the \textit{Students and Academics} group, indicating that both models achieve their most effective simplifications for these audiences. The \textit{Flesch-Kincaid Ease} and \textit{Grade} scores further confirm that the text is accessible to all user groups. While the Llama model outperforms in terms of \textit{BLEU} and \textit{SARI} scores, the \textit{GPT-4o} model achieves slightly higher \textit{FK Ease} scores and lower (i.e., better) \textit{FK Grade} scores compared to the Llama-3.1 model.
Overall, both models generate well-aligned, readable simplifications, effectively enhancing accessibility for a wide range of audiences.

\section{Conclusions}

Despite notable advances in large language models (LLMs) and their applications in text simplification and summarization, no running systems have yet been developed specifically for transforming text into plain language. In this paper, we introduced 
a web-based system designed to convert complex texts into plain language with a focus on enhancing accessibility and inclusion. 

In the future, we aim to expand the system's customization capabilities. Furthermore, we plan to leverage user feedback through reinforcement learning to further improve the quality and adaptability of text simplification.

\vspace{1em}
\noindent
\textbf{Acknowledgements.} 
This work was partially supported by the Saxon State Ministry for Science, Culture, and Tourism through the project ``Klartext.'' It was also partially supported by the German Federal Ministry of Education and Research (BMBF) as part of the Software Campus project ``LLM4Edu'' (grant number 01IS23070).

\bibliographystyle{splncs04}
 
\bibliography{ref}

\end{document}